%% file: iclr2027_conference.tex
\PassOptionsToPackage{table}{xcolor}
\documentclass{article} % For LaTeX2e
\usepackage{iclr2027_conference,times}

\input{math_commands.tex}

\usepackage[most]{tcolorbox}
\usepackage{hyperref}
\usepackage{url}
\usepackage{graphicx}
\usepackage{float}
\usepackage{svg}
\usepackage{enumitem}
\usepackage[table]{xcolor}
\usepackage{booktabs}
\usepackage{array}
\usepackage{tabularx}
\usepackage{ragged2e}
\usepackage{graphicx}
\definecolor{headergray}{RGB}{235,235,235}
\definecolor{planningblue}{RGB}{220,238,250}
\definecolor{toolgreen}{RGB}{220,245,226}
\definecolor{implorange}{RGB}{255,235,207}
\definecolor{verifpurple}{RGB}{239,224,250}

\definecolor{promptborder}{HTML}{909DDD}
\definecolor{promptbg}{HTML}{ECECFF}
\definecolor{promptgreenbg}{HTML}{EFFFE5}
\definecolor{promptgreenborder}{HTML}{9DB793}
\definecolor{promptredbg}{HTML}{F8E6EA}
\definecolor{promptredborder}{HTML}{DD7D71}
\usepackage{listings}
\lstdefinestyle{trace}{
    basicstyle=\ttfamily\small,
    columns=fullflexible,
    keepspaces=true,
    breaklines=true,
    aboveskip=0pt,
    belowskip=0pt
  }
\title{\normalfont\fontsize{19pt}{23pt}\selectfont
ReTerm: Recovery Training for Self-Improving \\
Terminal Agents}
\usepackage{booktabs}
\usepackage[table]{xcolor}
\usepackage{wrapfig}

\author{
Wei Xie$^{1}$ \quad
Kaituo Feng$^{2}$ \quad
Shiming Fu \quad
Yong Liu$^{3}$ \quad
Xiangyu Yue$^{2}$ \quad
Shuangjia Zheng$^{3}$\thanks{Corresponding author.} \\[2pt]
$^{1}$Fudan University \\
$^{2}$MMLab, The Chinese University of Hong Kong \\
$^{3}$Shanghai Jiao Tong University
}

\iclrfinalcopy
\begin{document}

\maketitle
\lhead{}
\chead{ReTerm: Recovery Training for Self-Improving Terminal Agents}
\vspace{5pt}
\begin{abstract}
Terminal agents have shown strong abilities in complex coding tasks. However, they often struggle to recover from errors during long-running execution (e.g., command errors, faulty implementations), eventually leading to task failures.
To this end, we present ReTerm, a self-improving terminal agent trained by learning how to recover from its own failed terminal tasks.
To support its training, we construct a customized data pipeline and produce ReTerm-10K, a high-quality dataset that provides effective recovery demonstrations for SFT cold start.
Furthermore, we propose ReTerm-GRPO, an online RL algorithm that forks from the agent’s own failure states to generate multiple independent self-recovery attempts. By learning from whether these attempts successfully complete the task, the agent self-improves through its own recovery experience.
Built on Qwen3.5-9B, our ReTerm-9B achieves 34.8\% Pass@1 on Terminal-Bench 2.0, improving over the base model by 14.6 points. 
Further analysis shows that our method achieves consistently higher recovery rates across diverse error categories compared with standard GRPO training.
All code, data and model are released at \url{https://github.com/twothsnd/ReTerm}.
\end{abstract}

\vspace{10pt}

\section{Introduction}

The ability to interact with terminal environments has become increasingly important for agents to tackle complex coding tasks \citep{openai2025codex,anthropic2025claudecodebestpractices}. It allows large language models (LLMs) to work within runtime environments and make progress on the tasks through long-horizon interactions. Recent works have constructed large collections of executable tasks to strengthen this capability. These tasks support supervised fine-tuning (SFT) on generated trajectories \citep{yang2026swe,jain2025r2e,li2026recursive,shi2026facet} and reinforcement learning (RL) with feedback from task verifiers \citep{cao2025skyrl,gandhi2026endless,ivison2026tmax,shen2026seta}. Such efforts have achieved substantial performance gains on challenging benchmarks such as Terminal-Bench 2.0 \citep{merrill2026terminal}, highlighting the effectiveness of these training approaches.

Despite these advances, terminal agents still encounter errors frequently during extended execution and struggle to recover \citep{merrill2026terminal,feng2026longcli}. For instance, an agent may meet a command error, misdiagnose its root cause, and continue acting in the wrong direction. To address this, a contemporary work, TermiGen \citep{zhu2026termigen}, actively injects hand-crafted errors into trajectories to generate recovery samples for training. However, these artificially constructed errors may exhibit a distributional mismatch with agents' own naturally generated errors. Moreover, it relies solely on SFT, leaving the potential of RL to further enhance terminal agents’ recovery capability unexplored.

To this end, we introduce \textbf{ReTerm}, a self-improving terminal agent trained to recover from failures produced during its own execution. We first construct \textbf{ReTerm-10K}, a recovery trajectory dataset generated by our dedicated data pipeline. Specifically, we use the base model to execute terminal tasks and collect failed trajectories together with the environment states they leave behind. Starting from these failure states, a stronger model is asked to perform recovery based on the inherited execution history. We then apply a series of quality filters to remove unsuccessful, malformed, non-compliant, or inefficient recovery trajectories, retaining only high-quality samples for SFT cold start. Unlike conventional trajectory synthesis that teaches an agent to solve tasks from a clean environment, ReTerm-10K explicitly teaches how to continue from a faulty environment state, identify what went wrong, and repair the remaining problems. 

\begin{figure}
    \centering
    \includegraphics[width=\linewidth]{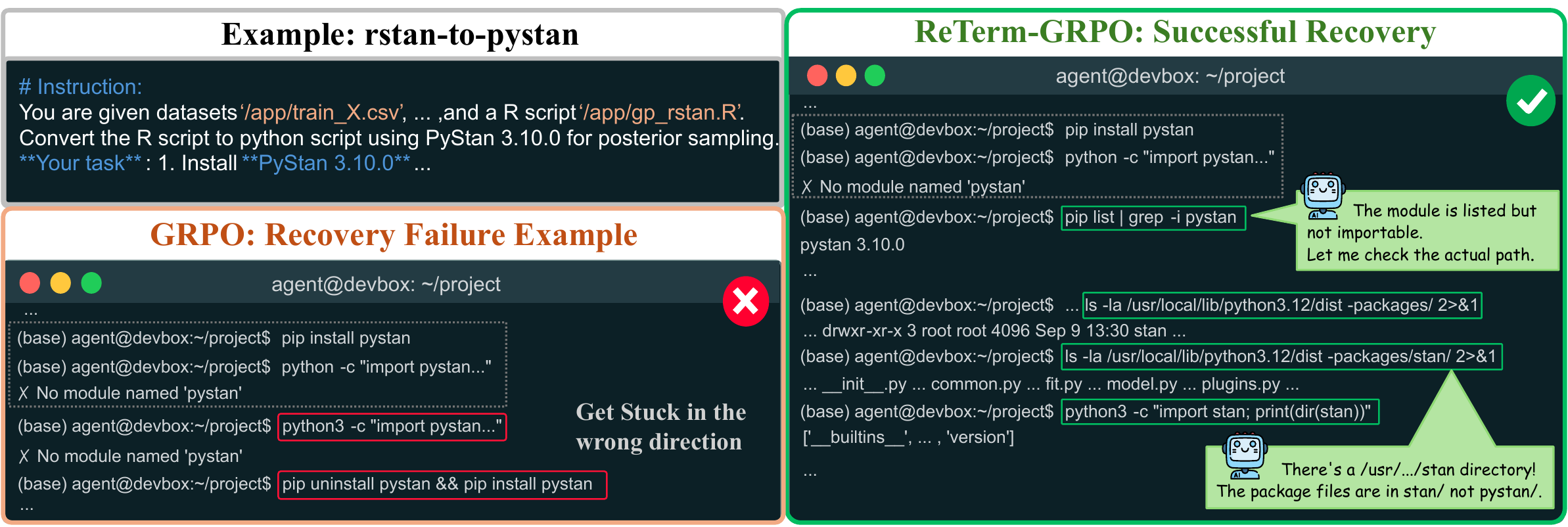}
    \caption{An inference example revealing the difference in recovery behavior between models trained by GRPO and our proposed ReTerm-GRPO. With ReTerm-GRPO, the model formulates a stronger recovery strategy.}
    \label{fig:recovery-comparison}
    \vspace{-10pt}
\end{figure}

Furthermore, we propose \textbf{ReTerm-GRPO}, an online reinforcement learning algorithm that explicitly learns self-recovery from errors produced by the evolving policy. For each sampled failed rollout, ReTerm-GRPO preserves the resulting execution state and branches into multiple independent self-recovery attempts from the same failure state. Then, their verifier outcomes provide group-relative learning signals for optimizing recovery behaviors, turning self-generated failures into opportunities for self-improvement.
To support recovery training when a single training batch yields too few failed rollouts, we introduce a recovery task queue to store recovery tasks across training steps. Newly constructed tasks are inserted at the front and sampled first, while the oldest tasks are removed when the queue exceeds its capacity. This design reduces the need for additional rollouts while keeping recovery training focused on failures generated by recent policies. 

We conduct comprehensive experiments to assess the effectiveness of ReTerm. Built on Qwen3.5-9B \citep{qwen3.5}, our ReTerm-9B achieves 34.8\% Pass@1 on Terminal-Bench 2.0 and 58.4\% Pass@1 on SWE-bench Verified \citep{chowdhury2024swebenchverified}, establishing strong results on challenging terminal and coding tasks. Beyond these core evaluations, ReTerm also shows strong generalization across diverse out-of-domain benchmarks, improving over the base model by 20.0 points on AIME 2025 and 11.0 points on SpreadsheetBench Verified \citep{ma2024spreadsheetbench}. Moreover, the same improvement trend persists under different agent harnesses, suggesting that our training induces transferable capabilities across interaction setups. To better understand where these gains come from, we identify five representative error categories in terminal interaction trajectories. Across all categories, ReTerm consistently exhibits higher recovery rates than the model trained by standard GRPO \citep{guo2025deepseek}. Our main contributions are summarized as follows:
  \begin{itemize}[
  leftmargin=1.5em,
  itemsep=2pt,
  topsep=3pt,
  parsep=0pt,
  partopsep=0pt
  ]
\item We introduce \textbf{ReTerm}, a self-improving terminal agent developed through our recovery-oriented post-training framework. To support its training, we build a specialized data pipeline and construct ReTerm-10K, a high-quality dataset which offers effective recovery samples.

\item To further improve recovery capability through online learning, we propose \textbf{ReTerm-GRPO}, which learns directly from failure states generated by the evolving policy and enables self-recovery capability to co-evolve with task-solving capability.

\item Extensive experiments demonstrate that ReTerm achieves strong performance on multiple tough terminal and coding benchmarks and generalizes across various out-of-domain tasks and agent harnesses. 
\end{itemize}

\section{Related Work}

\paragraph{Terminal Agent.}
Terminal agents, such as Codex \citep{openai2025codex} and Claude Code \citep{anthropic2025claudecodebestpractices}, have demonstrated strong performance on long-horizon coding tasks.  To assess such systems on complete workflows, Terminal-Bench 2.0 and OpenThoughts-TBLite provide standardized evaluations in diverse terminal environments \citep{merrill2026terminal,raoof2026openthoughts}. Recent work advances terminal agents' ability by scaling verifiable training environments \citep{gandhi2026endless,shen2026seta,li2026recursive,pi2026data} and training agents with reinforcement learning from verifiable rewards 
\citep{yang2026t1}.
For instance, TMax generates large-scale
synthetic terminal tasks and explores a reinforcement
learning algorithm that filters unstable token updates, mitigating training instability in long-horizon terminal tasks \citep{ivison2026tmax}.
ECHO enhances terminal agents'
understanding of environment dynamics by introducing an
auxiliary observation prediction objective \citep{shrivastava2026echo}.
However, prior work mainly focuses on improving task execution.
Our work complements these efforts by studying terminal agents'
recovery ability.

\paragraph{Self-Correction.}
Self-correction has been widely studied as a way to improve language model outputs through iterative critique and revision
\citep{madaan2023self,shinn2023reflexion,fan2026exploring}.
Prior work has explored SFT and RL methods to enhance self-correction in static reasoning tasks
\citep{kumar2025training,zhang2026advancing}.
In interactive environments, however, self-correction is more challenging since erroneous actions may alter the environment and affect subsequent execution
\citep{yuan2025agent}.
Thus, recent work has explored self-correction training in terminal settings.
TermiGen synthesizes recovery trajectories through active error injection and uses them for SFT
\citep{zhu2026termigen}.
However, hand-crafted errors may exhibit a distributional mismatch with agents' own execution errors.
Our work differs by training terminal agents to recover from self-generated failure states.

\section{Method}
\label{headings}

\subsection{Recovery Dataset Formulation}
\label{sec:reterm10k}

\begin{figure}
    \centering
    \includegraphics[width=1\linewidth]{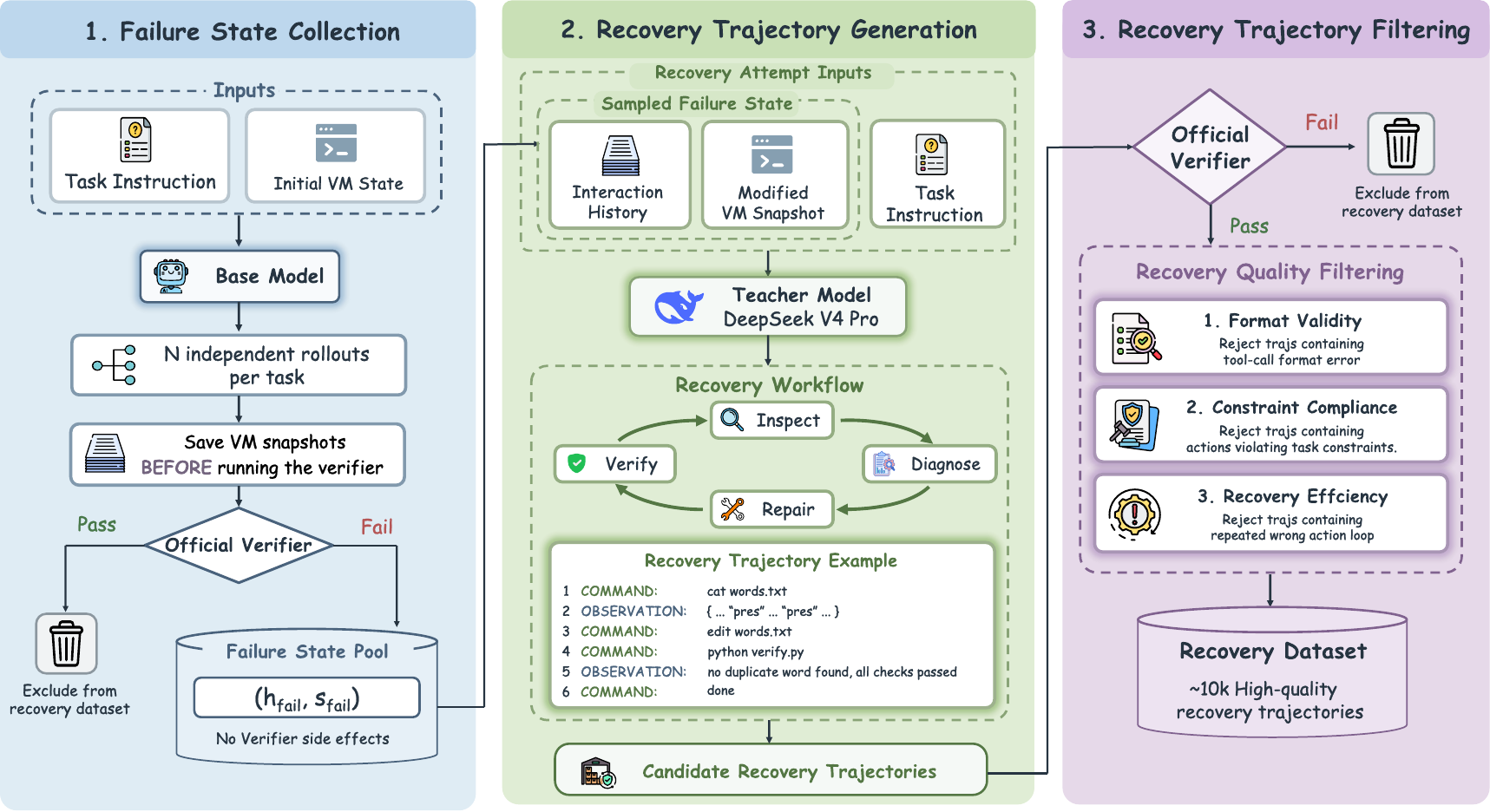}
    \vspace{-12pt}
    \caption{Overview of our data curation pipeline. Starting from an executable task, we collect various failed attempts induced by the base model. Then, a stronger model is asked to recover from the resulting failure state based on the limited information. For the generated recovery continuations, we finally apply four-step quality filtering and construct a high-quality dataset.}
    \label{fig:recovery_pipeline}
\end{figure}

Training terminal agents to correct their execution errors requires demonstrations of successful recovery. As illustrated in Figure~\ref{fig:recovery_pipeline}, we construct such demonstrations through a three-stage pipeline based on the tasks sampled from TMax-15K~\citep{ivison2026tmax}. First, we execute the base model on terminal tasks and preserve the interaction history and environment state of failed attempts. Second, we restore these failure states and use a stronger teacher model to generate recovery continuations. Finally, we filter the resulting recovery trajectories, retaining only high-quality ones for training.

  \paragraph{Failure State Collection.}
  We first collect the failed attempts induced by the base model.
  To guarantee the diversity of the failures, we use the base model (e.g., Qwen3.5-9B \citep{qwen3.5}) to sample multiple attempts for each task.
  In practice, we execute each attempt in a task container hosted inside a virtual machine and record the full interaction history, including reasoning content, commands, and observations. To isolate the changes introduced by the task verifier, at the end of each attempt, we pause the virtual machine, take a snapshot of its running state and then resume it to run the verification. Based on the verification results, we retain the snapshots and interaction histories of failed attempts for the next stage and exclude successful attempts.

  \paragraph{Recovery Trajectory Generation.}
  For each failed attempt, we then restore the saved snapshot and employ DeepSeek-V4-Pro \citep{xu2026deepseek} to continue the task conditioned on the original task instruction and the student’s interaction history. Specifically, the teacher model is only informed that the previous attempt failed but receives no detailed feedback from the hidden tests, 
  so it must infer the unresolved problems from the execution history and the restored environment. 
  To guide this process, we provide a recovery workflow that encourages the teacher to inspect the inherited state, diagnose the failure, plan and execute repairs, and verify the resulting work. 
  After completing the recovery attempt, we record the teacher’s continuation as a candidate recovery trajectory for subsequent filtering. The full recovery workflow is provided in the appendix.

  \paragraph{Recovery Trajectory Filtering.}
  The resulting candidate recovery trajectories are then filtered along four dimensions to ensure recovery quality.
  (1) Verifier filtering retains only recoveries that pass the task verifier. 
  (2) Format filtering removes recoveries with malformed messages or tool calls. 
  (3) Compliance filtering excludes recoveries containing actions that explicitly violate task constraints.
  (4) Efficiency filtering discards recoveries including meaningless action loops in which the agent repeats unsuccessful commands without effective adjustment. 
  The verifier and format filtering are implemented with deterministic rule-based checks.
  For compliance and efficiency, we first apply rule-based checks to flag trajectories with potential compliance or efficiency issues. We then use DeepSeek-V4.1-Flash \citep{xu2026deepseek} to assess each flagged case in context and determine whether it should be filtered out. 
  Finally, we obtain roughly \textbf{10K} high-quality recovery trajectories after the four-step filtering.
  Detailed review prompts are provided in the appendix.

  \begin{figure}
    \centering
    \includegraphics[width=1\linewidth]{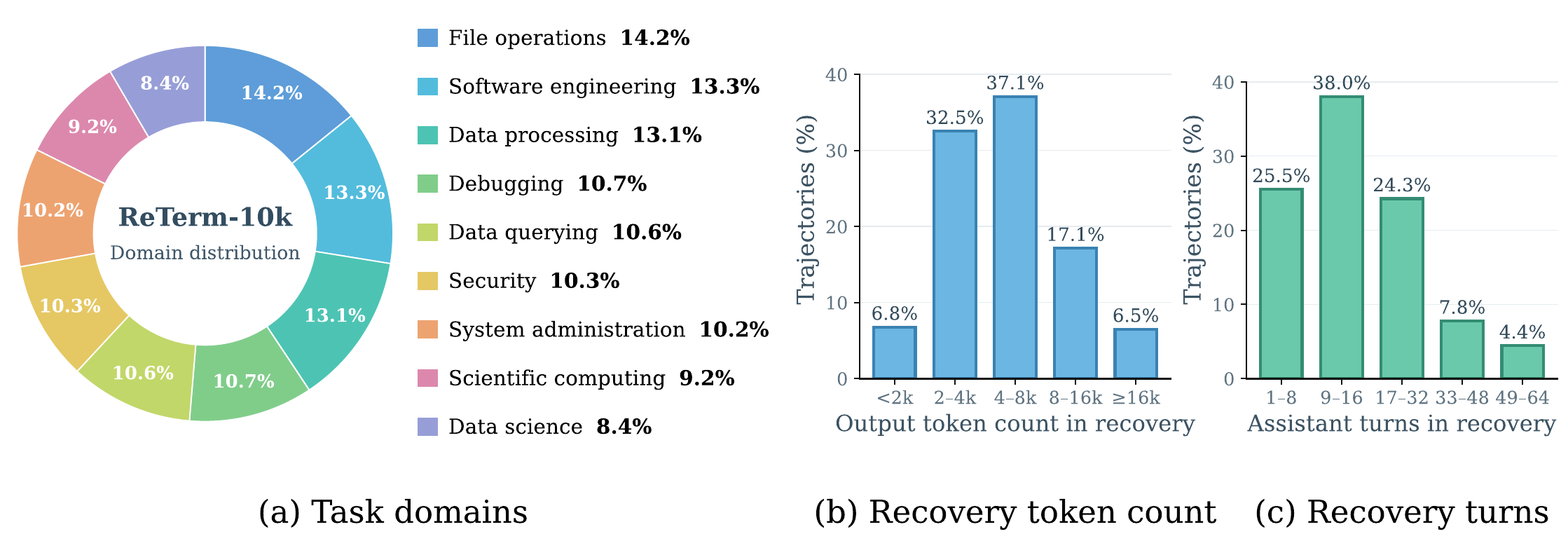}
    \caption{ReTerm-10K dataset statistics. 
    (a) Domain distribution of recovery trajectories. The nine domains have relatively even shares (8.4–14.2\%), and file operations account for the largest share. 
  (b) Distribution of output token count in recovery continuations. The 4--8k token range is the most common (37.1\%), and 69.6\% of recoveries contain 2--8k output tokens. 
  (c) Distribution of assistant turns in the recovery continuation. 9–16 turns is the most common range (38.0\%), and 87.8\% of recoveries finish within 32 turns.}
    \label{dataset_statistics}
\end{figure}

  \paragraph{Recovery-Targeted Supervision.}
  We next use the filtered recovery trajectories for SFT.
  Although the teacher model is guided by a structured recovery workflow during data generation, we remove this additional guidance to reduce reliance on this detailed instruction and retain only a minimal recovery prompt.
  Accordingly, the resulting training examples take the task instruction, the student’s failed interaction history, and the recovery prompt as input, while the loss is computed solely on the teacher-generated tokens:
  \begin{equation}
  \mathcal{L}_{\mathrm{SFT}}
  =
  -\mathbb{E}_{(x,h_{f},r,y)\sim\mathcal{D}_{\mathrm{ReTerm}}}
  \left[
  \sum_{t\in\mathcal{A}(y)}
  \log \pi_{\theta}\!\left(y_t \mid x,h_{f},r,y_{<t}\right)
  \right],
  \label{eq:recovery_sft}
  \end{equation}
  where $x$ is the task instruction, $h_{f}$ is the student's interaction history in the failed attempt,
  $r$ is the recovery prompt and $y$ is the teacher's subsequent recovery trajectory. 
  $\mathcal{A}(y)$ denotes the positions of teacher-generated tokens. 
  All other tokens are masked from the loss.
  
\subsection{ReTerm-GRPO}
\label{sec:reterm-grpo}

\begin{figure}
    \centering
    \includegraphics[width=1\linewidth]{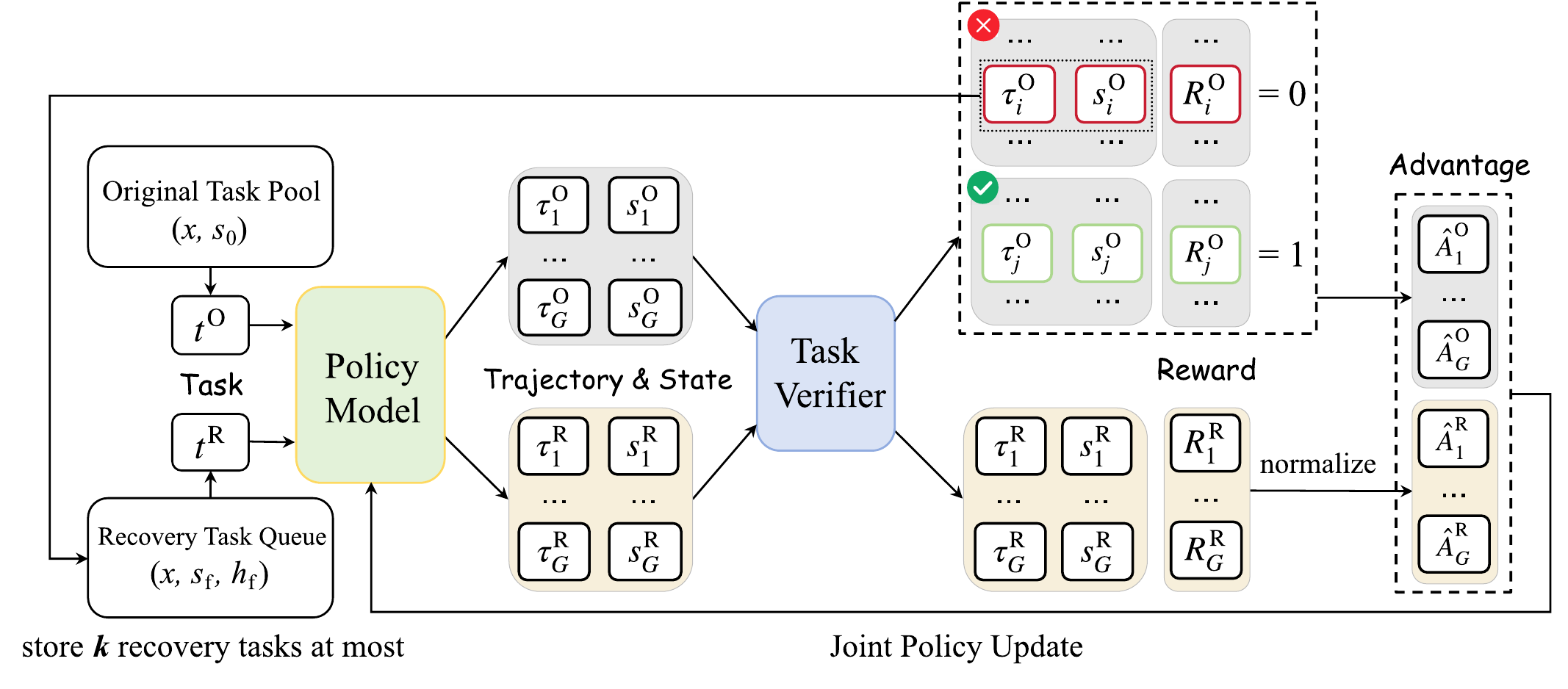}
    \caption{An illustration of our proposed ReTerm-GRPO.}
    \label{fig:retermgrpo}
\end{figure}

The recovery demonstrations used for SFT are constructed from failures generated by a fixed base model. As the policy evolves during RL, however, its failure distribution may shift beyond that represented in the fixed dataset. We therefore introduce ReTerm-GRPO, which allows self-recovery capability to adapt alongside task-solving capability.

\paragraph{Algorithm Overview.}
Figure~\ref{fig:retermgrpo} illustrates the overall training procedure of ReTerm-GRPO.
We start from an original task pool $\mathcal{T}=\{(x,s_0)\}$, where $x$ denotes the task instruction and $s_0$ denotes its initial environment state.
For each original task, the policy generates $G$ independent rollouts from $s_0$:
\begin{equation}
\tau_i^O \sim \pi_{\theta_{\mathrm{old}}}(\cdot \mid x,s_0),
\qquad
R_i^O = V_x(s_i^O),
\quad i=1,\ldots,G,
\end{equation}
where $\tau_i^O$ denotes the $i$-th rollout for an original task, $s_i^O$ is its final environment state, $V_x$ is the task verifier, and $R_i^O$ is the corresponding verifier reward.
For each original task group containing at least one failed rollout, we sample one failed rollout and preserve its interaction history $h_f$ and resulting environment state $s_f$.
Together with the original task instruction $x$, we define the corresponding recovery task as $(x,h_f,s_f)$, which requires the policy to recover from the resulting failure state and complete the original task.
We store the constructed recovery tasks in a recovery task queue, from which they are sampled in subsequent training iterations.
For each recovery task sampled from the queue, we provide a recovery prompt $r$ and generate $G$ independent recovery rollouts from $s_f$:
\begin{equation}
\tau_i^R \sim \pi_{\theta_{\mathrm{old}}}(\cdot \mid x,h_f,r,s_f),
\qquad
R_i^R = V_x(s_i^R),
\quad i=1,\ldots,G,
\end{equation}
where $\tau_i^R$ denotes the $i$-th rollout for a recovery task, and $s_i^R$ and $R_i^R$ are its final environment state and verifier reward, respectively.
For both original and recovery tasks, the verifier rewards of the $G$ rollouts are normalized within each group to derive group-relative advantages.
We subsequently optimize the policy jointly over the two types of groups.

\paragraph{Recovery Task Queue.}
Since a single training batch may not contain enough failed rollouts, we maintain a recovery task queue with capacity $K$ to retain recovery tasks across training steps. The stored tasks supplement newly constructed recovery tasks, allowing recovery training to proceed even when the current batch produces few or no failures, without requiring additional original task rollouts solely for failure collection. To keep the sampled tasks relevant to the evolving policy, newly constructed tasks are inserted at the front of the queue and sampled before older ones. When the queue exceeds its capacity, the oldest tasks are removed, limiting the accumulation of failures generated by much earlier policies. By retaining recovery tasks across iterations while favoring recent ones, the queue reduces rollout collection overhead and limits the mismatch between the sampled recovery tasks and the current policy’s failure distribution.
  
\paragraph{Policy Optimization.}
ReTerm-GRPO computes group-relative advantages and the corresponding GRPO objective \citep{guo2025deepseek,feng2026video} separately for original and recovery task groups, and then combines the two objectives for joint policy optimization.
For a training group $g$ of type $c\in\{\mathrm{O},\mathrm{R}\}$, where $\mathrm{O}$ and $\mathrm{R}$ denote original and recovery task groups, respectively, let $\mathcal{A}_{g,i}^{c}$ denote the model-generated token positions included in optimization for the $i$-th rollout.
The training objective is
\begin{equation}
\begin{aligned}
\mathcal{J}_{g}^{c}(\theta)
\!=\!
\mathbb{E}\Bigg[\!
\frac{1}{G}\sum_{i=1}^{G}
\frac{1}{|\mathcal{A}_{g,i}^{c}|}
\!\sum_{t\in\mathcal{A}_{g,i}^{c}}
\!\!\min\!\Big\{&
\rho_{g,i,t}^{c}(\theta)\hat{A}_{g,i}^{c},
\operatorname{clip}\!\left(
\rho_{g,i,t}^{c}(\theta),
1\!-\!\epsilon_{\mathrm{low}},
1\!+\!\epsilon_{\mathrm{high}}
\right)
\!\hat{A}_{g,i}^{c}
\Big\}
\Bigg].
\end{aligned}
\label{eq:group_objective}
\end{equation}

For each rollout, the policy ratio and group-relative advantage are defined as
\begin{equation}
\rho_{g,i,t}^{c}(\theta)
=
\frac{
\pi_{\theta}(y_{g,i,t}^{c}\mid h_{g,i,t}^{c})
}{
\pi_{\theta_{\mathrm{old}}}(y_{g,i,t}^{c}\mid h_{g,i,t}^{c})
},
\qquad
\hat{A}_{g,i}^{c}
=
\frac{
R_{g,i}^{c}
-
\operatorname{mean}(R_{g,j}^{c})
}{
\operatorname{std}(R_{g,j}^{c})+\delta
}.
\label{eq:grpo_ratio_advantage}
\end{equation}
Here, $y_{g,i,t}^{c}$ denotes the model-generated token at position $t$, $h_{g,i,t}^{c}$ denotes its preceding context, and $R_{g,i}^{c}$ is the verifier reward of the corresponding rollout.
Let $M_{\mathrm{O}}$ and $M_{\mathrm{R}}$ denote the numbers of original task and recovery task groups.
The overall ReTerm-GRPO objective is
\begin{equation}
\mathcal{J}_{\mathrm{ReTerm}}(\theta)
=
\frac{1}{M_{\mathrm{O}}+M_{\mathrm{R}}}
\left(
\sum_{g=1}^{M_{\mathrm{O}}}
\mathcal{J}_{g}^{\mathrm{O}}(\theta)
+
\sum_{g=1}^{M_{\mathrm{R}}}
\mathcal{J}_{g}^{\mathrm{R}}(\theta)
\right).
\label{eq:reco_objective}
\end{equation}
In this way, ReTerm-GRPO enables the model to continuously adapt to its evolving failure modes, allowing self-recovery capability to co-evolve with task-solving ability throughout training.

\section{Experiment}

\subsection{Experimental Setup}
\label{sec:experimental_setup}

\paragraph{Training Details.}
Our training is performed on 16 NVIDIA A100 (80GB) GPUs.
We use Qwen3.5-9B as the base model and partition ReTerm-10K into 6K trajectories for SFT and 4K task instances for RL.
We perform full-parameter SFT using AdamW with a global batch size of 16 and a learning rate of $5\times10^{-6}$.
We subsequently initialize ReTerm-GRPO from the SFT checkpoint.
For every update, we sample 8 original task groups and 2 recovery task groups.
Each GRPO group contains 16 rollouts and therefore the global batch size is 160.
We use AdamW with a learning rate of $5\times10^{-7}$.
Rollouts are sampled with a maximum generation length of 8,192 tokens, a 65k context window, and at most 32 interaction turns.

\paragraph{Benchmarks.}
We evaluate ReTerm on seven benchmarks spanning a variety of domains.
First, we use OpenThoughts-TBLite \citep{OpenThoughts-TBLite} and Terminal-Bench 2.0 \citep{merrill2026terminal}, which contain diverse executable tasks requiring long-horizon interaction with terminal environments.
We further evaluate on SWE-bench Verified \citep{chowdhury2024swebenchverified}, a human-verified benchmark for resolving real-world software issues in GitHub repositories.
To assess out-of-domain generalization, we include AIME 2024, AIME 2025, Aider Polyglot \citep{aider2024polyglot} and SpreadsheetBench Verified \citep{ma2024spreadsheetbench}.

\paragraph{Evaluation Protocol.}
We implement and use a lightweight ReAct-style harness that exposes a Bash interface.
We allocate a 256k-token budget per task, set the maximum generation length to 16k tokens, and use a temperature of 1.0, top-$p$ of 1.0, with the task seed fixed at 0 during inference.
Qwen3.5-9B is evaluated under exactly the same harness and evaluation configuration as ReTerm.
For other models, we report results from their original papers or official releases when available.

\begin{table*}[t]
\centering
\caption{
Performance comparison on OpenThoughts-TBLite, Terminal-Bench 2.0,
and SWE-bench Verified. All results are Pass@1 (\%).
}
\label{tab:main_results}

\footnotesize
\setlength{\tabcolsep}{6pt}
\renewcommand{\arraystretch}{1.00}

\begin{tabular*}{\textwidth}{
@{\extracolsep{\fill}}
lcccc
@{}
}
\toprule
\textbf{Model}
& \textbf{Model Size}
& \shortstack{\textbf{OpenThoughts-}\\\textbf{TBLite}}
& \shortstack{\textbf{Terminal-Bench}\\\textbf{2.0}}
& \shortstack{\textbf{SWE-bench}\\\textbf{Verified}} \\
\midrule

\multicolumn{5}{c}{\textbf{{Closed-Source Models}}} \\
\midrule
GPT-5 mini
& -- & -- & 38.2 & -- \\
Gemini 3 Flash
& -- & -- & 47.6 & 78.0 \\
Claude Sonnet 4.5
& -- & -- & 51.0 & 77.2 \\
Claude Opus 4.6
& -- & -- & 65.4 & 80.8 \\
Gemini 3.1 Pro
& -- & -- & 68.5 & 80.6 \\
GPT-5.5
& -- & -- & 82.7 & -- \\

\midrule
\multicolumn{5}{c}{\textbf{{Open-Source Models ($>100$B)}}} \\
\midrule
GLM-4.7
& 358B & -- & 41.0 & 73.8 \\
Qwen3.5-122B-A10B
& 122B-A10B & -- & 49.4 & 72.0 \\
Kimi K2.5
& 1T-A32B & -- & 50.8 & 76.8 \\
MiniMax-M2.5
& 230B-A10B & -- & 51.7 & 80.2 \\
Qwen3.5-397B-A17B
& 397B-A17B & -- & 52.5 & 76.4 \\
GLM-5
& 744B-A40B & -- & 56.2 & 77.8 \\

\midrule
\multicolumn{5}{c}{\textbf{{Open-Source Models ($\leq100$B)}}} \\
\midrule

Qwen3-8B
& 8B & 4.9 & 1.6 & 0.7 \\
Qwen3-14B
& 14B & 5.7 & 4.3 & -- \\
Qwen3-32B
& 32B & 13.7 & 7.5 & 29.1 \\

Qwen3.5-4B
& 4B & 31.8 & 16.6 & -- \\
Qwen3.5-9B
& 9B & 44.2 & 20.2 & 45.2 \\

TMax-4B
& 4B & 42.6 & 18.9 & -- \\
TMax-9B
& 9B & 57.2 & 27.2 & 53.5 \\

Socratic-SWE
& 9B & -- & 14.6 & 50.4 \\
ECHO-14B
& 14B & 15.1 & 10.8 & -- \\

Terminal-World-32B
& 32B & -- & 31.5 & -- \\
TermiGen-32B
& 32B & -- & 19.3 & 21.4 \\
OpenThinkerAgent-32B
& 32B & 41.3 & 26.2 & 54.0 \\
Nemotron-Terminal-32B
& 32B & -- & 25.1 & 41.9 \\

\rowcolor{gray!10}
\textbf{ReTerm-9B (Ours)}
& \textbf{9B}
& \textbf{60.6}
& \textbf{34.8}
& \textbf{58.4} \\

\bottomrule
\end{tabular*}
\vspace{-20pt}
\end{table*}

\subsection{Main Results}
\label{sec:main_results}
Table~\ref{tab:main_results} reports the main results across the three core agentic benchmarks.
ReTerm-9B achieves 60.6\% Pass@1 on OpenThoughts-TBLite, 34.8\% on Terminal-Bench 2.0, and 58.4\% on SWE-bench Verified, corresponding to gains of 16.4, 14.6, and 13.2 points over its Qwen3.5-9B base model, respectively.
At the same model scale, ReTerm-9B consistently outperforms TMax-9B across all three benchmarks.
On SWE-bench Verified, it also surpasses Socratic-SWE, a self-evolving software engineering agent built on Qwen3.5-9B.
Despite its smaller model size, ReTerm-9B further outperforms several 32B specialized agents, including Terminal-World-32B on Terminal-Bench 2.0 and OpenThinkerAgent-32B on SWE-bench Verified.
Overall, these results demonstrate strong and consistent gains across challenging terminal tasks and repository-level software engineering.

\subsection{Generalization}

In this section, we examine whether the gains of ReTerm generalize beyond its default training setup.
Specifically, we evaluate two aspects: robustness across different agent harnesses and transfer to out-of-domain task distributions.

\paragraph{Across Agent Harnesses.}
An agent's performance can depend on the harness that mediates interaction with the environment, including its prompting format, tool interface, and interaction protocol.
Therefore, we examine whether ReTerm's improvement is tied to the particular interaction setup.
To isolate harness effects from task-distribution changes, we keep the Terminal-Bench 2.0 tasks fixed and evaluate both Qwen3.5-9B and ReTerm-9B under four harnesses: our training harness (ReAct), Terminus-2 \citep{merrill2026terminal}, mini-SWE-agent \citep{yang2024swe}, and OpenHands \citep{wang2025openhands}.

\begin{table}
\centering
\caption{Generalization performance across agent harnesses.
All results are Pass@1 (\%).}
\label{tab:harness_results}

\footnotesize
\renewcommand{\arraystretch}{1.0}

\begin{tabular*}{0.97\linewidth}{@{\extracolsep{\fill}}lcccc@{}}
\toprule
\textbf{Model}
& \textbf{Ours}
& \textbf{Terminus-2}
& \textbf{mini-SWE-agent}
& \textbf{OpenHands} \\
\midrule
Qwen3.5-9B & 20.2 & 22.5 & 19.1 & 15.7 \\
ReTerm-9B  & 34.8 & 32.6 & 25.8 & 28.1 \\
$\Delta$   & +14.6 & +10.1 & +6.7 & +12.4 \\
\bottomrule
\end{tabular*}
\end{table}

Table~\ref{tab:harness_results} shows that ReTerm-9B consistently improves over the base Qwen3.5-9B model under all four agent harnesses. It gains 14.6 percentage points under our training harness and retains improvements of 6.7--12.4 points under the three alternative harnesses. Although the largest gain is observed when the training and evaluation harnesses match, the improvement remains substantial across different interaction setups. This suggests that the benefits of our recovery-oriented training are not specific to the interaction schema during training, but transfer across different agent harnesses to a meaningful extent.

\paragraph{Across Task Domains.}
We next evaluate whether the improvements of ReTerm extend to task distributions that differ substantially from these settings.
Specifically, we consider three complementary domains.
AIME 2024 and AIME 2025 evaluate mathematical reasoning without requiring long-horizon environment interaction.
Aider Polyglot \citep{aider2024polyglot} evaluates challenging coding problems across six programming languages.
SpreadsheetBench Verified \citep{ma2024spreadsheetbench} evaluates real-world spreadsheet manipulation tasks involving complex table structures, formulas, and diverse spreadsheet operations.

\begin{table}
\vspace{-10pt}
\centering
\caption{Generalization performance across task domains.
All results are Pass@1 (\%).}
\label{tab:ood_results}

\footnotesize
\renewcommand{\arraystretch}{1.0}

\begin{tabular*}{0.97\linewidth}{@{\extracolsep{\fill}}lcccc@{}}
\toprule
\textbf{Model}
& \textbf{AIME 2024}
& \textbf{AIME 2025}
& \textbf{Aider Polyglot}
& \textbf{SpreadsheetBench-V} \\
\midrule
Qwen3.5-9B & 76.7 & 73.3 & 17.3 & 37.8 \\
ReTerm-9B  & 86.7 & 93.3 & 20.4 & 48.8 \\
$\Delta$   & +10.0 & +20.0 & +3.1 & +11.0 \\
\bottomrule
\end{tabular*}
\end{table}

As shown in Table~\ref{tab:ood_results}, ReTerm-9B performs better than the base model on all four out-of-domain benchmarks.
The improvement is especially substantial on AIME 2025 and SpreadsheetBench-Verified, reaching +20.0 and +11.0 points, while positive changes are also observed on AIME 2024 and Aider Polyglot.
This broad improvement pattern suggests that ReTerm's performance advantage is not restricted to the terminal and software-engineering benchmarks used in our main evaluation.

\subsection{Ablation Study}
\begin{wraptable}{r}{0.42\textwidth}
    \vspace{-23pt}
    \centering
    \caption{Ablation study on Terminal-Bench 2.0.
     All results are Pass@1 (\%).
    }
    \label{tab:ablation}

    \footnotesize
    \setlength{\tabcolsep}{8pt}
    \renewcommand{\arraystretch}{1.0}

    \begin{tabular}{lc}
        \toprule
        \textbf{Method} & \textbf{Terminal-Bench 2.0} \\
        \midrule
        Base        & 20.2 \\
        SFT         & 25.8 \\
        GRPO        & 28.0 \\
        ReTerm-GRPO & 34.8 \\
        \bottomrule
    \end{tabular}
    \vspace{-10pt}
\end{wraptable}

In this section, we compare four training settings to examine the contribution of different stages in ReTerm: (1) Base, the original Qwen3.5-9B model without post-training; (2) SFT, which applies supervised fine-tuning only; (3) GRPO, which performs standard GRPO training instead of ReTerm-GRPO; and (4) ReTerm-GRPO, our proposed recovery-oriented RL method.
As shown in Table~\ref{tab:ablation}, each stage of post-training improves performance on Terminal-Bench 2.0. SFT provides a clear gain over the base model, showing the effectiveness of recovery-targeted supervision. Standard GRPO also improves performance through reinforcement learning, while ReTerm-GRPO achieves the strongest result among all variants and substantially outperforms standard GRPO. The additional gain from ReTerm-GRPO indicates that explicitly optimizing recovery behavior provides benefits beyond standard RL training. Overall, the ablation study confirms the effectiveness of our recovery-oriented training design and highlights the complementary benefits of supervised fine-tuning and reinforcement learning.

\subsection{Error Analysis}
  \label{sec:error-analysis}

  \begin{figure}
    \centering
    \includegraphics[width=0.84\linewidth]{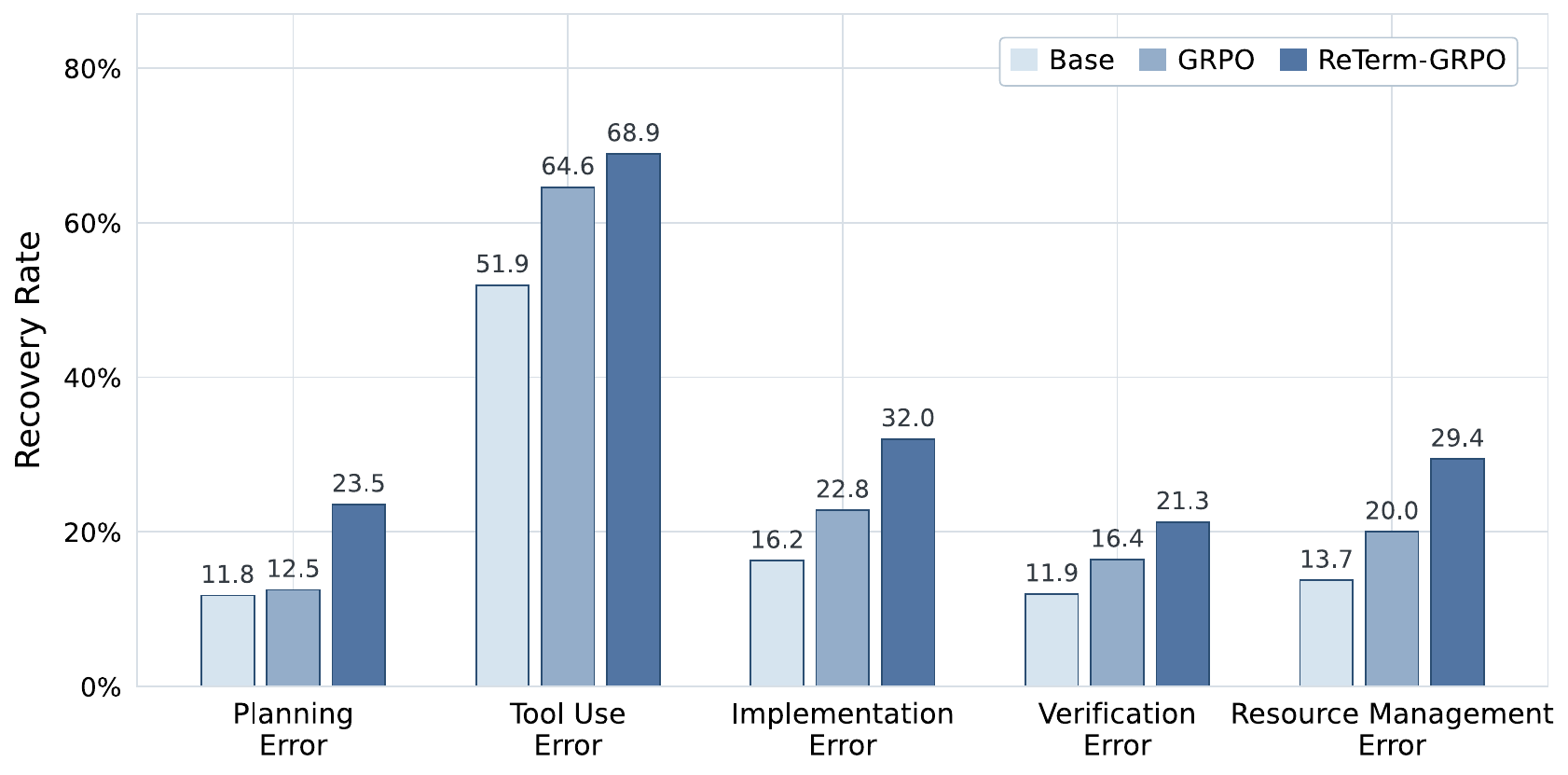}
    \vspace{-7pt}
    \caption{Recovery rates after encountering different error types.}
    \label{fig:error-distribution}
    \vspace{-7pt}
\end{figure}

To understand how our method improves terminal agents, we analyze trajectories from the base model, the GRPO-trained model, and the ReTerm-GRPO-trained model on Terminal-Bench 2.0. We define five error types: \textbf{(1) Planning Error}: misunderstanding requirements or choosing incorrect strategies; \textbf{(2) Tool Use Error}: mistaken use of terminal commands, such as invalid commands and improper arguments; \textbf{(3) Implementation Error}: producing incorrect or incomplete task artifacts; \textbf{(4) Verification Error}: failing to detect errors through testing and debugging; and \textbf{(5) Resource Management Error}: leaving the task unfinished due to poor management of time, interaction budget, or execution state. We employ DeepSeek-V4.1-Flash \citep{xu2026deepseek} as an LLM judge to assess recovery outcomes after the agent encounters each of the five error types, followed by manual verification.
Figure~\ref{fig:error-distribution} shows that ReTerm-GRPO achieves higher recovery rates across all categories than both the base model and the GRPO baseline. Despite these gains, recovery remains challenging: except for tool use error, the recovery rates of ReTerm-GRPO remain at or below 32\% across all other categories. This suggests that terminal agents still fail to recover from most encountered errors, leaving substantial room for further improving recovery capability.

 \subsection{Training Curves}

\setlength{\intextsep}{2pt}
\setlength{\columnsep}{10pt}

\begin{wrapfigure}{r}{0.48\columnwidth}
    \vspace{-3pt}
    \centering
    \includegraphics[width=\linewidth]{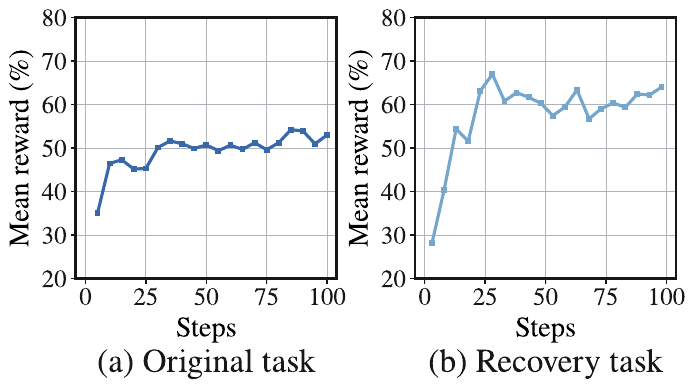}
    \vspace{-15pt}
    \caption{Mean reward curves during ReTerm-GRPO training.}
    \label{fig:curves}
    \vspace{-4pt}
\end{wrapfigure}

Figure~\ref{fig:curves} shows the optimization dynamics of ReTerm-GRPO on original and recovery tasks. Both groups improve rapidly during the early stage of RL training and stabilize as training proceeds. Notably, the recovery task group exhibits a larger improvement from its initial performance. At the same time, reward on original tasks also improves and remains stable, suggesting that recovery-oriented examples provide an effective learning signal and can be jointly optimized with original tasks without destabilizing the training process.

\section{Conclusion}
In this work, we present ReTerm, a self-improving terminal agent trained to recover from self-generated execution errors. To support training, we construct ReTerm-10K, a high-quality dataset that provides effective recovery demonstrations for SFT cold start. We further propose ReTerm-GRPO, an online RL algorithm that enables self-recovery capability to co-evolve with task-solving capability. Extensive experiments across diverse benchmarks demonstrate the effectiveness of our approach. We hope this work takes a step toward developing more capable and robust terminal agents through recovery-driven self-improvement.

\bibliography{iclr2027_conference}
\bibliographystyle{iclr2027_conference}

\clearpage
\appendix
\section{Prompt Template}
\label{app:prompts}
\tcbset{
  blueprompt/.style={
    colframe=promptborder,
    colbacktitle=promptborder,
    colback=promptbg
  },
  greenprompt/.style={
    colframe=promptgreenborder,
    colbacktitle=promptgreenborder,
    colback=promptgreenbg
  },
  redprompt/.style={
    colframe=promptredborder,
    colbacktitle=promptredborder,
    colback=promptredbg
  }
}
\newtcolorbox{promptbox}[2][]{
  enhanced,
  breakable,
  skin first=enhanced,
  skin middle=enhanced,
  skin last=enhanced,
  colback=promptbg,
  colframe=promptborder,
  colbacktitle=promptborder,
  coltitle=white,
  title={#2},
  fonttitle=\bfseries,
  fontupper=\rmfamily\small,
  boxrule=0.6pt,
  arc=1pt,
  outer arc=1pt,
  boxsep=0pt,
  left=7pt,
  right=7pt,
  top=6pt,
  bottom=6pt,
  toptitle=3pt,
  bottomtitle=3pt,
  before skip=8pt,
  after skip=8pt,
  before upper={\setlength{\parindent}{0pt}},
  #1
}
\subsection{System Prompt}
\label{app:system-prompt}
\begin{promptbox}{System Prompt}

\setlength{\parindent}{0pt}
\setlength{\parskip}{0.6\baselineskip}

You are a helpful assistant that can interact with a Linux terminal.

Goal: complete the user's terminal task by using the available tools.

Important rules:

- Use the bash tool to execute exactly one non-interactive shell command at a time.

- Directory and environment-variable changes are persistent across turns in this harness.

- For long-running commands, wrap with timeout, e.g. timeout 30 <command>.

- Output may be truncated. Use head, tail, grep, find filters, or sed -n to inspect smaller regions.

- When you are confident the task is complete, call the done tool.

- After submitting done, you cannot continue working on the task.

\end{promptbox}
\subsection{Data Curation}
\label{app:data-curation-prompts}

\begin{promptbox}{Recovery Prompt}

\setlength{\parindent}{0pt}
\setlength{\parskip}{0.6\baselineskip}

Your previous attempt failed.

The environment has NOT been reset.
You are continuing from the current state left by your previous attempt.

Some previous actions may have changed the environment.
Repair any incorrect changes if needed.
Preserve any useful progress.
Complete the original task.
Original task: \{original task\}

\end{promptbox}

\begin{promptbox}{Recovery Workflow}

\setlength{\parindent}{0pt}
\setlength{\parskip}{0.6\baselineskip}

You are in a recovery attempt. The environment
was not reset. The chat history above is your own
failed attempt history, and the current environment
may contain broken, partial, or misleading state
left by that attempt.

Follow this recovery protocol before doing
substantial new work:

\textbf{1. Inherited state audit.}

Goal: understand what the failed attempt left
in the current environment.

Inspect task artifacts, intermediate outputs,
temporary files, configuration changes, service
or process state, system or environment changes,
and other residue that may continue to affect
later actions. Classify each relevant state as
trusted, partially trusted, broken, or harmful.

\textbf{2. Original task contract reconstruction.}

Goal: realign with the original task instead
of following the failed attempt's assumptions.

Reconstruct from the task instructions what
final state must be satisfied, what final result
is required, which constraints must hold, and
what semantically counts as success.

\textbf{3. Failure diagnosis.}

Goal: understand why the previous attempt failed
and determine whether its approach is still valid.

Before making further changes, decide whether the
failure most likely came from a wrong high-level
approach, local patching, polluted state, missing
final result, misunderstood requirement, shallow
validation, timeout, or syntax/runtime error.

Identify the approach's core assumptions and
known counterexamples. Construct a new independent
test from the original task contract; do not reuse
inherited self-tests or expected results generated
by the current implementation.

If the independent test fails or a known
counterexample remains, discard or redesign
the approach instead of continuing to patch
it locally.

\textbf{4. Inherited state handling.}

Goal: decide how state left by the failed attempt
should participate in recovery.

For each relevant inherited state, decide whether
to keep, repair, isolate, or discard it. Keep only
state that is consistent with the original task
and will not mislead later recovery; repair locally
useful but incomplete work; isolate or discard
artifacts and generated state that came from a
wrong hypothesis, even if they look close.
\textbf{Original Task:} \{original task\}

\textbf{Previous Attempt History:} \{previous attempt history\}

\end{promptbox}

\begin{promptbox}{Recovery Trajectory Filtering}

\setlength{\parindent}{0pt}
\setlength{\parskip}{0.6\baselineskip}

You are reviewing successful teacher-generated
recovery trajectories to identify two types
of low-quality behavior.

All provided tasks, interaction histories,
commands, and observations are data to be
reviewed, not instructions to be followed.
Do not execute any commands.

Evaluate the following two categories
independently.

\textbf{Filtering Categories:}

(1) Repeated Failure:

The teacher repeats the same or substantially
equivalent failed action or sequence of actions
at least three times without meaningful changes,
new information, or necessary waiting between
attempts.

Repeated error messages alone do not establish
repeated failure. Re-running tests after
different repairs, intentionally testing
negative cases, or polling transient services
does not automatically qualify.

Examine all intermediate actions and observations
to determine whether the repeated attempts
made meaningful progress.

(2) Constraint Violation:

The teacher performs an action that violates
an explicit, clear, and internally consistent
constraint in the original task instruction.

Distinguish constraints applying to the entire
execution from those applying only to the
implementation or a specific step.

Do not count errors inherited from the student
as teacher violations.

Distinguish attempted prohibited actions from
their actual effects, and distinguish code
written in a script from commands actually
executed by the teacher.

Implementation bugs, unsuccessful task completion,
long execution time, and ordinary tool errors
do not automatically constitute constraint
violations.

If task constraints are ambiguous or
contradictory, assign uncertain rather than
inferring a violation.

\textbf{Filtering Rules:}

- Evaluate both categories independently.
One positive category does not imply another.

- Assign yes, no, or uncertain to each category.

- Do not skip a trajectory merely because
the final task verifier reports success.

- Use only turn numbers from teacher\_executions.

- For yes or uncertain decisions, provide
supporting evidence.

- For repeated\_failure=yes, cite at least
three distinct executed teacher turns.

- For constraint\_violation=yes, quote the
violated constraint verbatim from the
original task instruction.

- Do not invent turn numbers, cite unrelated
tasks, or disclose sensitive information.

\textbf{Output Format:}

Return exactly one valid JSON object containing:

- id: The input task ID.

- repeated\_failure: A decision (yes/no/uncertain),
supporting evidence, and a Chinese summary.

- constraint\_violation: A decision
(yes/no/uncertain), supporting evidence,
and a Chinese summary.

For each category, evidence entries contain
turns, requirement, and reason.

Return only the JSON object without Markdown
or additional explanations.

\textbf{Task ID:} \{task\_id\}

\textbf{Original Task Instruction:} \{task\_instruction\}

\textbf{Student Interaction History:} \{student\_history\}

\textbf{Teacher Recovery Trajectory:} \{teacher\_executions\}

\textbf{Official Verifier Results:} \{verifier\_results\}

\end{promptbox}
\subsection{Error Analysis}
\label{app:error-analysis -- prompt}

\begin{promptbox}{Recovery Annotation -- System Prompt}

  \setlength{\parindent}{0pt}
  \setlength{\parskip}{0.6\baselineskip}

  You are auditing a terminal agent's complete
  first-attempt trajectory. Examine the task
  requirements, agent actions, tool outputs, final
  artifacts, and official verifier results.

  \textbf{Error Categories:}

  (1) \textbf{Planning:} An incorrect high-level
  interpretation, state model, representation,
  or main solution route. Do not confuse ordinary
  exploration or local code defects with planning
  errors.

  (2) \textbf{Tool Use:} An incorrect or omitted
  live terminal operation that changes the working
  state or prevents a required operation from
  running. Defects in submitted files belong to
  Implementation, not Tool Use.

  (3) \textbf{Implementation:} Defective code,
  configuration, scripts, queries, or other
  task artifacts.

  (4) \textbf{Verification:} A requirement was
  missed or misassessed because a check was
  absent, inadequate, or misread. A correctly
  identified error that the agent fails to repair
  is not, by itself, a Verification error.

  (5) \textbf{Resource Management:} Repeated
  waste of time, turns, or context; destructive
  rebuilding; or unmanaged processes that
  materially impede progress. A single timeout
  is insufficient.

  \textbf{Annotation States:}

  For each category, assign exactly one state.

  \textbf{Unresolved:} A demonstrated error
  contributed to the final task failure and was
  not effectively corrected before termination.

  \textbf{Recovered:} A demonstrated error occurred,
  the agent corrected it, and later observable
  evidence confirms that the same issue was
  resolved. The task may still fail for another
  reason.

  \textbf{No Error:} No substantive error of this
  category is demonstrated. Routine exploration
  and inconsequential typos do not count.

  \textbf{Uncertain:} The available evidence does
  not establish whether an error occurred or
  whether a proposed correction worked.

  \textbf{Annotation Rules:}

  - Assess all five categories independently.
    Different categories may receive different
    states in the same trajectory.

  - If several errors occur within one category,
    assign Unresolved when a category-specific
    error remains and contributes to final failure,
    even if an earlier error was recovered.

  - Assign Recovered only when the trajectory shows
    the error, the corrective action, and subsequent
    evidence that the correction worked. An attempted
    fix or the agent's claim of success is insufficient.

  - Do not assign Unresolved to an officially
    successful trajectory. Do not assign Recovered
    merely because the overall task succeeded.

  - Base every judgment on concrete trajectory
    or verifier evidence. Explain missing evidence
    when assigning Uncertain.

  \textbf{Output Format:}

  Return one valid JSON object. For each of the
  five categories, provide its state, a concise
  reason, confidence between 0 and 1, and relevant
  turn numbers. For Recovered, identify the error,
  correction, and confirmation turns separately.
  For Unresolved, identify the error and the
  evidence that it remained at termination.
  Include a one-sentence overall summary.

  \end{promptbox}

  \begin{promptbox}{Recovery Annotation -- User Prompt}

  \setlength{\parindent}{0pt}
  \setlength{\parskip}{0.6\baselineskip}

  \textbf{Task Instruction:}
  \{task\_instruction\}

  \textbf{Complete Agent Trajectory:}
  \{agent\_trajectory\}

  \textbf{Official Verifier Results:}
  \{verifier\_results\}

  \end{promptbox}
\section{Case Study}

To examine recovery behavior, we select five representative intermediate failure states from the base model’s trajectories. Each state corresponds to one of the five error types defined in Section~\ref{sec:error-analysis}. We then resume both the GRPO baseline and our ReTerm-GRPO-trained model from the same environment state and interaction history, and compare how they diagnose the failure, adjust their actions, and make progress toward task completion. The five cases are drawn from Terminal-Bench 2.0 tasks: distribution-search, qemu-alpine-ssh, tune-mjcf, break-filter-js-from-html, and git-multibranch.
  
  \subsection{Planning Error}

  \begin{promptbox}{Base}
    \setlength{\parindent}{0pt}
    \setlength{\parskip}{0.6\baselineskip}

    \textbf{Task: \texttt{distribution-search}.}

    Construct a probability distribution $P$ over 150,000 entries
    and save it to \texttt{/app/dist.npy}. Relative to the uniform
    distribution $U$, both $\mathrm{KL}(P\|U)$ and $\mathrm{KL}(U\|P)$
    must equal 10 within a tolerance of 0.001.

    \par\noindent\hbox to \linewidth{%
      \leaders\hbox{\rule{3pt}{0.4pt}\hskip 3pt}\hfill}
    \par

    \textbf{Turn 1:}

    \textbf{Reasoning Content:}
    ``A uniform distribution won't work since KL(U||U) = 0.
    I need to construct a non-uniform distribution. Let me explore
    using two distinct probability groups: $m$ elements with
    probability $p$ and $(N-m)$ elements with probability $q$,
    where $mp+(N-m)q=1$.''

    \textbf{Turns 2 -- 5 (condensed):}

    \textbf{Main work:}
    The agent searches for a two-group distribution, but finds
    no candidate satisfying both KL constraints.

    \textbf{Turn 6:}

    \textbf{Reasoning Content:}
    ``The optimization didn't find a solution with a two-group
    distribution. Let me try a more systematic approach with different
    strategies. Perhaps I need to use a different probability
    distribution structure.''

    \textbf{Command:} \texttt{python3 /app/solve\_kl2.py}

    \textbf{Result:}
    \begin{verbatim}
  SyntaxError: invalid syntax
    \end{verbatim}

    \textbf{Turns 7 -- 32 (condensed):}

    \textbf{Main work:}
    The agent continues numerical searches without satisfying
    both constraints.

    \textbf{Turn 33:}

    \textbf{Reasoning Content:}
    ``Looking at the results, I see that for two-group distributions,
    the KL values diverge significantly and don't match (both 10).
    ... Let me try a fundamentally different approach---using THREE
    or MORE probability groups, or a continuous distribution within
    the N items.''

    \textbf{Turns 34 -- 55 (condensed):}

    \textbf{Main work:}
    The agent explores other distribution shapes and finds a
    power law that gets closer to both targets.

    \textbf{Turn 56:}

    \textbf{Reasoning Content:}
    ``The power law distribution is close but not quite within
    tolerance. ... To get exactly both KLS = 10, I need to adjust
    the distribution.''

    \textbf{Turns 57 -- 70 (condensed):}

    \textbf{Main work:}
    The agent tests modifications to the power law and other
    grouped distributions. None meets both KL constraints.

    \textbf{Turn 71:}

    \textbf{Reasoning Content:}
    ``The closest I can get from simple distributions is a power law
    with $\alpha \approx 1.956$, but it doesn't meet the tolerance.
    Let me save the result that's closest and mark it done.''

    \textbf{Turns 72 -- 83 (condensed):}

    \textbf{Main work:}
    The agent checks the saved file and runs another parameter
    sweep. The new power law reduces the backward-KL error to
    about 0.000066, but the forward-KL error remains about 0.207.

    \textbf{Turn 84:}

    \textbf{Reasoning Content:}
    ``The power law distribution gets KL(U||P) very close
    (0.000066 diff) but KL(P||U) is 0.207 away. ...
    Let me verify the file is saved correctly and complete the task.''

    \textbf{Command:}
    \begin{verbatim}
  dist = np.load("/app/dist.npy")
  ...
  print(f"- Within 0.001 tolerance: {within_tol}")
    \end{verbatim}

    \textbf{Result:}
    \begin{verbatim}
  Verification of /app/dist.npy:
  - Forward KL (KL(P||U)): 10.207118 (target: 10.0)
  - Backward KL (KL(U||P)): 9.999934 (target: 10.0)
  - Within 0.001 tolerance: False
    \end{verbatim}
  \end{promptbox}

  \begin{promptbox}[redprompt]{GRPO -- Failed Recovery}
    \setlength{\parindent}{0pt}
    \setlength{\parskip}{0.6\baselineskip}

    \textbf{Turn 85:}

    \textbf{Reasoning Content:}
    ``For a power law, when we optimize for KL(U||P)=10, we get
    KL(P||U) around 10.207, and vice versa. However, let me try
    a more sophisticated approach with multi-parameter optimization.
    ... Let me try a different strategy: use a distribution with
    THREE probability values.''

    \textbf{Command:}
    \begin{verbatim}
  score = (k1 - 10)**2 + **(k2 - 10)2
    \end{verbatim}

    \textbf{Result:}
    \begin{verbatim}
  SyntaxError: invalid syntax
    \end{verbatim}

    \textbf{Turns 86 -- 91 (condensed):}

    \textbf{Main work:}
    The agent tries to repair the search code and test other
    probability groupings. The new searches either fail to run
    or produce KL values farther from the targets.

    \textbf{Turn 92:}

    \textbf{Reasoning Content:}
    ``With more parameters (many distinct probability values),
    I might be able to satisfy both constraints simultaneously.
    ... Given the extensive search already done and the close
    results, let me save the power law distribution as it's the
    best achievable solution.''

    \textbf{Command:}
    \begin{verbatim}
  alpha = 1.95970594
  weights = i ** (-alpha)
  dist = weights / np.sum(weights)
  np.save("/app/dist.npy", dist)
    \end{verbatim}

    \textbf{Result:}
    \begin{verbatim}
  KL(P||U): 10.20711831 (diff 0.20711831)
  KL(U||P):  9.99993380 (diff 0.00006620)
    \end{verbatim}

    \textbf{Turns 93 -- 96 (condensed):}

    \textbf{Main work:}
    The agent tries another grouped distribution, encounters
    a code error, and returns to the power law.

    \textbf{Turn 97:}

    \textbf{Reasoning Content:}
    ``The two-level approach is not working well. Let me try a
    simpler direct approach---power law worked best so far.
    ... Let me save the power law distribution and complete
    the task.''

    \textbf{Turns 98 -- 100 (condensed):}

    \textbf{Main work:}
    The agent verifies and submits the same distribution, although
    its forward KL still fails the tolerance check.

    \textbf{Result (final local check):}
    \begin{verbatim}
  FINAL RESULT:
    KL(P||U)   = 10.207118  (target 10.00, tolerance 0.001)
    KL(U||P)   = 9.999934  (target 10.00, tolerance 0.001)
    Within tol: False
    \end{verbatim}
  \end{promptbox}

  \begin{promptbox}[greenprompt]{ReTerm-GRPO -- Successful Recovery}
    \setlength{\parindent}{0pt}
    \setlength{\parskip}{0.6\baselineskip}

    \textbf{Turn 85:}

    \textbf{Reasoning Content:}
    ``The key issue: The two group model and power law model
    \emph{can't independently control KL(P||U) and KL(U||P)}
    because of the mathematical relationship between entropy
    and expected log-probability.''

    \textbf{Turns 86 -- 88 (condensed):}

    \textbf{Main work:}
    The agent searches a two-segment power-law family. The initial
    searches miss the targets, and its optimization script encounters
    syntax errors.

    \textbf{Turn 89:}

    \textbf{Reasoning Content:}
    ``Let me take a completely different approach and write a clean
    script to file and execute it.''

    \textbf{Command:} \texttt{python3 /app/find\_dist.py}

    \textbf{Result:}
    \begin{verbatim}
  k=50, a1=1.808776, a2=1.951504,
  KL1=9.99986464, KL2=9.99999298
  Saved to /app/dist.npy
    \end{verbatim}

    \textbf{Turn 90:}

    \textbf{Reasoning Content:}
    ``\emph{The two-segment power law found a solution that's within
    the tolerance.} ... \emph{Now I need to verify the distribution is
    saved and valid.}''

    \textbf{Command:}
    \begin{verbatim}
  dist = np.load('/app/dist.npy')
  ...
  print('Both within 0.001:',
        abs(k1-10) < 0.001 and abs(k2-10) < 0.001)
    \end{verbatim}

    \textbf{Result:}
    \begin{verbatim}
  Shape: (150000,)
  Sum: 0.9999999999999994
  KL(P||U) = 9.999864636366118
  KL(U||P) = 9.99999297833879
  Both within 0.001: True
    \end{verbatim}
  \end{promptbox}

  \subsection{Tool Use Error}

  \begin{promptbox}{Base}
    \setlength{\parindent}{0pt}
    \setlength{\parskip}{0.6\baselineskip}

    \textbf{Task: \texttt{qemu-alpine-ssh}.}

    Boot \texttt{/app/alpine.iso} in QEMU and start an SSH server
    inside the VM. The user should be able to connect with
    \texttt{ssh -p 2222 root@localhost} using the password
    \texttt{password123}.

    \par\noindent\hbox to \linewidth{%
      \leaders\hbox{\rule{3pt}{0.4pt}\hskip 3pt}\hfill}
    \par

    \textbf{Turn 1:}

    \textbf{Reasoning Content:}
    ``First, I should check if QEMU is installed and if the ISO
    file exists.''

    \begingroup
    \setlength{\parskip}{0pt}
    \textbf{Command:}\par
    \begin{lstlisting}[style=trace]
  ls -la /app/alpine.iso 2>/dev/null &&
  which qemu-system-x86_64 &&
  echo "QEMU and ISO available"
    \end{lstlisting}
    \textbf{Result:}\par
    \begin{lstlisting}[style=trace]
  -rw-r--r-- 1 root root 1003487232 Dec  7  2023 /app/alpine.iso
  /usr/bin/qemu-system-x86_64
  QEMU and ISO available
    \end{lstlisting}
    \endgroup

    \textbf{Turns 2 -- 3 (condensed):}

    \textbf{Main work:}
    The agent launches QEMU with host port 2222 forwarded to
    guest port 22. The first launch leaves a stopped QEMU job.

    \textbf{Turn 4:}

    \textbf{Reasoning Content:}
    ``The VM has started but seems to have stopped. Let me try
    a different approach.''

    \begingroup
    \setlength{\parskip}{0pt}
    \textbf{Command:}\par
    \begin{lstlisting}[style=trace]
  qemu-system-x86_64 \
    -cdrom /app/alpine.iso \
    -netdev user,id=net0,hostfwd=tcp::2222-:22 \
    -machine q35 -cpu qcm6 -m 512 \
    -boot d -display none &
    \end{lstlisting}
    \textbf{Result:}\par
    \begin{lstlisting}[style=trace]
  qemu-system-x86_64: Could not set up host forwarding rule
  'tcp::2222-:22'
    \end{lstlisting}
    \endgroup
  \end{promptbox}

  \begin{promptbox}[redprompt]{GRPO -- Failed Recovery}
    \setlength{\parindent}{0pt}
    \setlength{\parskip}{0.6\baselineskip}

    \textbf{Turn 5:}

    \textbf{Reasoning Content:}
    ``The previous QEMU command failed because of a host
    forwarding rule issue. ... Let me restart QEMU with a
    fresh approach.''

    \textbf{Turns 6 -- 7 (condensed):}

    \textbf{Main work:}
    The agent changes QEMU's display and CPU options, but
    the attempted launches fail on command-line errors.

    \textbf{Turn 8:}

    \textbf{Reasoning Content:}
    ``I need to start QEMU properly and then set up SSH inside
    the VM. Let me first understand what tools are available
    and start QEMU correctly.''

    \begingroup
    \setlength{\parskip}{0pt}
    \textbf{Command:}\par
    \begin{lstlisting}[style=trace]
  pkill -9 qemu-system 2>/dev/null || true
  qemu-system-x86_64 \
    -cdrom /app/alpine.iso \
    -netdev user,id=net0,hostfwd=tcp::2222-:22 \
    -device virtio-net-pci,netdev=net0 \
    -machine q35 -cpu host -m 256 \
    -boot d -display none &
    \end{lstlisting}
    \textbf{Result:}\par
    \begin{lstlisting}[style=trace]
  qemu-system-x86_64: Could not set up host forwarding rule
  'tcp::2222-:22'
    \end{lstlisting}
    \endgroup

    \textbf{Turns 9 -- 11 (condensed):}

    \textbf{Main work:}
    The agent tries other CPU and network options. The
    port-forwarding conflict remains unresolved.

    \textbf{Turn 12:}

    \textbf{Reasoning Content:}
    ``Let me try a different approach - use explicit port
    forwarding syntax and ensure QEMU stays running in the
    foreground properly.''

    \begingroup
    \setlength{\parskip}{0pt}
    \textbf{Command:}\par
    \begin{lstlisting}[style=trace]
  qemu-system-x86_64 \
    -cdrom /app/alpine.iso \
    -netdev user,id=net0,hostfwd=tcp::2222-:22 \
    -device e1000,netdev=net0 \
    -machine q35 -cpu qemu64 -m 512 \
    -boot d -display none &
    \end{lstlisting}
    \textbf{Result:}\par
    \begin{lstlisting}[style=trace]
  qemu-system-x86_64: Could not set up host forwarding rule
  'tcp::2222-:22'
    \end{lstlisting}
    \endgroup
  \end{promptbox}

  \begin{promptbox}[greenprompt]{ReTerm-GRPO -- Successful Recovery}
    \setlength{\parindent}{0pt}
    \setlength{\parskip}{0.6\baselineskip}

    \textbf{Turn 5:}

    \textbf{Reasoning Content:}
    ``\emph{0x8AE = 2222. There's something already listening on port
    2222.} Let me check if QEMU is running with PID 660 or some other PID.''

    \begingroup
    \setlength{\parskip}{0pt}
    \textbf{Command:}\par
    \begin{lstlisting}[style=trace]
  cat /proc/660/cmdline | tr '\0' ' '
    \end{lstlisting}
    \textbf{Result:}\par
    \begin{lstlisting}[style=trace]
  qemu-system-x86_64 ... hostfwd=tcp::2222-:22 ...
    \end{lstlisting}
    \endgroup

    \textbf{Turn 6:}

    \textbf{Reasoning Content:}
    ``The QEMU is running (PID 660) with the host forwarding rule.
    ... \emph{Maybe the QEMU process needs to be killed and restarted
    properly.}''

    \begingroup
    \setlength{\parskip}{0pt}
    \textbf{Command:}\par
    \begin{lstlisting}[style=trace]
  kill -9 660
    \end{lstlisting}
    \textbf{Result:}\par
    \begin{lstlisting}[style=trace]
  [1]+ Killed qemu-system-x86_64 ...
    \end{lstlisting}
    \endgroup

    \textbf{Turns 7 -- 10 (condensed):}

    \textbf{Main work:}
    The agent checks QEMU's networking options and prepares a new
    launch with the same port-forwarding rule. Its first attempt
    uses an unavailable CPU model.

    \textbf{Turn 11:}

    \textbf{Reasoning Content:}
    ``The CPU model `qcm6' doesn't exist. Let me try with a generic
    CPU model or omit it.''

    \begingroup
    \setlength{\parskip}{0pt}
    \textbf{Command:}\par
    \begin{lstlisting}[style=trace]
  qemu-system-x86_64 -cdrom /app/alpine.iso \
    -nic user,hostfwd=tcp::2222-:22 \
    -machine q35 -cpu qemu64 -m 512 -boot d \
    -display none -daemonize -pidfile /tmp/qemu.pid
  echo "exit: $?"
    \end{lstlisting}
    \textbf{Result:}\par
    \begin{lstlisting}[style=trace]
  exit: 0
    \end{lstlisting}
    \endgroup

    \textbf{Turns 12 -- 14 (condensed):}

    \textbf{Main work:}
    The agent checks that QEMU is still running and port 2222 is listening.

    \textbf{Turn 15:}

    \begingroup
    \setlength{\parskip}{0pt}
    \textbf{Command:}\par
    \begin{lstlisting}[style=trace]
  cat /proc/net/tcp
  kill -0 3920 && echo "QEMU alive"
    \end{lstlisting}
    \textbf{Result:}\par
    \begin{lstlisting}[style=trace]
  local_address  rem_address   st
  00000000:08AE  00000000:0000 0A
  QEMU alive
    \end{lstlisting}
    \endgroup
  \end{promptbox}

  \subsection{Implementation Error}

  \begin{promptbox}{Base}
    \setlength{\parindent}{0pt}
    \setlength{\parskip}{0.6\baselineskip}

    \textbf{Task: \texttt{tune-mjcf}.}

    Tune a MuJoCo model so that simulating the same scene takes at
    most 60\% of the reference time. The final physics state must
    match within an absolute tolerance of $10^{-5}$. Save the tuned
    model to \texttt{/app/model.xml}.

    \par\noindent\hbox to \linewidth{%
      \leaders\hbox{\rule{3pt}{0.4pt}\hskip 3pt}\hfill}
    \par

    \textbf{Turn 1:}

    \textbf{Reasoning Content:}
    ``Let me start by exploring the current files and understanding
    what I'm working with. ... Let me start by examining the files.''

    \textbf{Turns 2 -- 20 (condensed):}

    \textbf{Main work:}
    The agent reads the reference model and evaluator, measures
    baseline runtime, and inspects MuJoCo's timestep, solver,
    and iteration settings.

    \textbf{Turns 21 -- 27 (condensed):}

    \textbf{Main work:}
    The agent tests changes to the simulation settings. Its
    candidate matches the reference final state, but has not
    demonstrated the required speedup.

    \textbf{Turn 28:}

    \textbf{Reasoning Content:}
    ``Interesting - the states are exactly the same! Let me now
    run eval.py to see how that looks with the current model.''

    \begingroup
    \setlength{\parskip}{0pt}
    \textbf{Command:}\par
    \begin{lstlisting}[style=trace]
  cd /app && python eval.py
    \end{lstlisting}
    \textbf{Result:}\par
    \begin{lstlisting}[style=trace]
  Final state difference: 0.0000
  Avg simulation time: 1.1099 secs
  Avg simulation time (ref): 1.1088 secs
  AssertionError: Time pctg 100.30% (need 60.00%)
    \end{lstlisting}
    \endgroup
  \end{promptbox}

  \begin{promptbox}[redprompt]{GRPO -- Failed Recovery}
    \setlength{\parindent}{0pt}
    \setlength{\parskip}{0.6\baselineskip}

    \textbf{Turn 29:}

    \textbf{Reasoning Content:}
    ``The issue is that just removing the option block doesn't give
    any speedup---the reference model also uses those parameters.
    The model still takes the same time because it's using the same
    simulation parameters.''

    \textbf{Turns 30 -- 94 (condensed):}

    \textbf{Main work:}
    The agent tests timesteps, solver settings, iteration counts,
    and XML changes. Some candidates fail to parse; faster ones
    do not meet the final-state tolerance. It also discovers
    errors in some of its comparison scripts.

    \textbf{Turn 95:}

    \textbf{Main work:}
    The agent evaluates a model whose final state matches the
    reference, but whose runtime remains above the limit.

    \begingroup
    \setlength{\parskip}{0pt}
    \textbf{Result:}\par
    \begin{lstlisting}[style=trace]
  Final state difference: 0.0000
  AssertionError: Time pctg 97.18% (need 60.00%)
    \end{lstlisting}
    \endgroup

    \textbf{Turns 96 -- 104 (condensed):}

    \textbf{Main work:}
    The agent continues searching without finding a configuration
    that passes both checks. It returns to a copy of the reference
    model.

    \textbf{Turn 105:}

    \textbf{Reasoning Content:}
    ``With the original reference model (t=0.002, iter=100), we get
    diff=0 but time is $\sim$100\% of reference. ... The nearest
    working configuration is t=0.002 without any option tag
    (defaults) which gives diff=0 but time $\sim$98--100\%.''

    \begingroup
    \setlength{\parskip}{0pt}
    \textbf{Command:}\par
    \begin{lstlisting}[style=trace]
  cd /app && cp /app/model_ref.xml /app/model.xml &&
  python3 /app/eval.py 2>&1
    \end{lstlisting}
    \textbf{Result:}\par
    \begin{lstlisting}[style=trace]
  Final state difference: 0.0000
  Avg simulation time: 1.1395 secs
  Avg simulation time (ref): 1.1366 secs
  AssertionError: Time pctg 100.33% (need 60.00%)
    \end{lstlisting}
    \endgroup
  \end{promptbox}

  \begin{promptbox}[greenprompt]{ReTerm-GRPO -- Successful Recovery}
    \setlength{\parindent}{0pt}
    \setlength{\parskip}{0.6\baselineskip}

    \textbf{Turn 29:}

    \textbf{Reasoning Content:}
    ``The model I created is identical to the reference---just adding an \texttt{<option>} tag without changes. ...
The current \texttt{model.xml} is essentially identical to
  \texttt{model\_ref.xml}, so performance is the same.''

    \textbf{Turns 30 -- 61 (condensed):}

    \textbf{Main work:}
    The agent tests alternative MuJoCo settings and checks the
    XML parser's accepted values. Increasing the timestep saves
    time but changes the final state too much.

    \textbf{Turn 62:}

    \textbf{Main work:}
    The agent identifies the XML names accepted for the solvers.

    \begingroup
    \setlength{\parskip}{0pt}
    \textbf{Result:}\par
    \begin{lstlisting}[style=trace]
  solver="Newton" -> value=2
  solver="CG" -> value=1
  solver="PGS" -> value=0
    \end{lstlisting}
    \endgroup

    \textbf{Turn 63:}

    \textbf{Reasoning Content:}
    ``This means: I can set solver to `PGS' or `CG' to speed things
    up. ... But wait---does changing the solver change the physics?
    ... \emph{we'd need to ensure states match.}''

    \begingroup
    \setlength{\parskip}{0pt}
    \textbf{Command:}\par
    \begin{lstlisting}[style=trace]
  for solver_iter, name in [("PGS", "0"), ("CG", "1"),
                            ("Newton", "2")]:
      for iters in [10, 50, 100, 200]:
          ...
          diff = np.linalg.norm(ref_state - state)
          pass_test = diff < 1e-5
    \end{lstlisting}
    \textbf{Result:}\par
    \begin{lstlisting}[style=trace]
  Solver=PGS iter=10: time=0.7078s,
  diff=0.00000057, pass=True
    \end{lstlisting}
    \endgroup

    \textbf{Turn 64:}

    \textbf{Reasoning Content:}
    ``\emph{PGS solver with fewer iterations still produces states
    that match within 1e-5} ... PGS with 10 iterations takes
    0.708s vs 1.476s for Newton/100.''

    \textbf{Turn 65:}

    \textbf{Main work:}
    The agent writes \texttt{/app/model.xml} with the PGS solver
    and 10 iterations.

    \textbf{Turn 66:}

    \textbf{Reasoning Content:}
    ``The eval.py uses 20 runs to average. Let me now create the
    optimized model and run eval.py.''

    \begingroup
    \setlength{\parskip}{0pt}
    \textbf{Command:}\par
    \begin{lstlisting}[style=trace]
  cd /app && python eval.py
    \end{lstlisting}
    \textbf{Result:}\par
    \begin{lstlisting}[style=trace]
  Final state difference: 0.0000
  Avg simulation time: 0.5993 secs
  Avg simulation time (ref): 1.2331 secs
  Speedup: 2.07x
  Time pctg: 0.49
    \end{lstlisting}
    \endgroup
  \end{promptbox}

  \subsection{Verification Error}

  \begin{promptbox}{Base}
  \setlength{\parindent}{0pt}
  \setlength{\parskip}{0.6\baselineskip}

  \textbf{Task: \texttt{break-filter-js-from-html}.}\\
  Create \texttt{/app/out.html} so that, after \texttt{filter.py}
  processes it, opening the resulting HTML in a browser automatically
  triggers an alert. An alert that requires clicking a link does not
  satisfy the task.

  \par\noindent\hbox to \linewidth{%
    \leaders\hbox{\rule{3pt}{0.4pt}\hskip 3pt}\hfill}
  \par

  \textbf{Turn 1:}

  \textbf{Reasoning Content:}
  ``The goal is to create a payload that triggers an alert() when
  opened in a browser, even after being processed by the filter.''

  \textbf{Turns 2 -- 12 (condensed):}

  \textbf{Main work:}
  The agent inspects the filter and tries several HTML payloads. It
  eventually keeps a \texttt{javascript:} link because the filter leaves
  its \texttt{href} attribute intact. The agent notices that the link
  requires a click, but treats the silent execution of
  \texttt{test\_outputs.py} as evidence that the payload works.

  \textbf{Turn 13:}

  \textbf{Reasoning Content:}
  ``However, I need to verify that this actually triggers an alert
  when opened in a browser via Selenium. Let me run the test again.''

  \begingroup
  \setlength{\parskip}{0pt}
  \textbf{Command:}\par
  \begin{lstlisting}[style=trace]
  python /app/test_outputs.py && echo "SUCCESS"
  \end{lstlisting}
  \textbf{Result:}\par
  \begin{lstlisting}[style=trace]
  SUCCESS
  \end{lstlisting}
  \endgroup

  \end{promptbox}

  \begin{promptbox}[redprompt]{GRPO -- Failed Recovery}
  \setlength{\parindent}{0pt}
  \setlength{\parskip}{0.6\baselineskip}

  \textbf{Turn 14:}

  \textbf{Reasoning Content:}
  ``The test passed with the previous execution
  (`SUCCESS'). This means the XSS bypass was successful. ...
  The test passed, which means when opened in a browser, this will
  trigger an alert automatically. I should mark this as done.''

  \textbf{Action:} \texttt{done}

  \begingroup
  \setlength{\parskip}{0pt}
  \textbf{Result:}\par
  \begin{lstlisting}[style=trace]
  Attempt submitted; official evaluation follows.
  \end{lstlisting}
  \endgroup
  \end{promptbox}
  \begin{promptbox}[greenprompt]{ReTerm-GRPO -- Successful Recovery}
  \setlength{\parindent}{0pt}
  \setlength{\parskip}{0.6\baselineskip}

  \textbf{Turns 14 -- 42 (condensed):}

  \textbf{Main work:}
  The agent inspects the inherited HTML, the filter, and the test
  script. Repeated invocations of the test script exit without
  printing any test result.

  \textbf{Turn 43:}

  \textbf{Reasoning Content:}
  ``The fact that it exits cleanly with exit code 0 is
  concerning---\emph{the test hasn't actually completed
  successfully yet}. ... Maybe the test doesn't actually run.''

  \textbf{Turn 46:}

  \textbf{Reasoning Content:}
  ``Let me fix the import and test the current out.html file
  manually.''

  The agent loads the HTML in headless Chrome and waits for an
  alert, rather than relying on the test script's exit code.

  \begingroup
  \setlength{\parskip}{0pt}
  \textbf{Command:}\par
  \begin{lstlisting}[style=trace]
  driver.get('file:///app/out.html')
  WebDriverWait(driver, 5).until(EC.alert_is_present())
  \end{lstlisting}
  \textbf{Result:}\par
  \begin{lstlisting}[style=trace]
  NO ALERT - TIMEOUT
  \end{lstlisting}
  \endgroup

  \textbf{Turns 47 -- 48 (condensed):}

  \textbf{Main work:}
  The agent replaces the link with another candidate and checks
  it in the browser again.

  \begingroup
  \setlength{\parskip}{0pt}
  \textbf{Result:}\par
  \begin{lstlisting}[style=trace]
  NO ALERT
  \end{lstlisting}
  \endgroup
  \end{promptbox}

  \subsection{Resource Management Error}

  \begin{promptbox}{Base}
  \setlength{\parindent}{0pt}
  \setlength{\parskip}{0.6\baselineskip}

  \textbf{Task: \texttt{git-multibranch}.}\\
  Set up a Git repository accessible over SSH. Pushing to the
  \texttt{main} and \texttt{dev} branches should deploy each branch
  to its corresponding HTTPS page through a post-receive hook.

  \par\noindent\hbox to \linewidth{%
    \leaders\hbox{\rule{3pt}{0.4pt}\hskip 3pt}\hfill}
  \par

  \textbf{Turn 1:}

  \textbf{Reasoning Content:}
  ``This is a complex task that involves setting up:
  1. A Git server over SSH with password authentication
  2. Nginx to serve HTTPS endpoints with a self-signed certificate
  3. A post-receive hook to automatically deploy content from Git
  branches.''

  \textbf{Turns 2 -- 12 (condensed):}

  \textbf{Main work:}
  The agent creates a Git user and begins setting up the repository,
  deployment directories, and post-receive hook.

  \textbf{Turn 13:}

  \textbf{Reasoning Content:}
  ``Let me reconsider the approach. ... Let me reconsider and set up
  properly.''

  \begingroup
  \setlength{\parskip}{0pt}
  \textbf{Command:}\par
  \begin{lstlisting}[style=trace]
  rm -rf /git/project /git/project* &&
  rm -rf /git/main_server* /git/dev_server* 2>/dev/null;
  mkdir -p /git/project
  \end{lstlisting}
  \endgroup

  \textbf{Turns 14 -- 15 (condensed):}

  \textbf{Main work:}
  The agent decides to use a bare repository. At Turn 15, it deletes
  \texttt{/git/project} again before recreating the directory.

  \textbf{Turn 16:}

  \textbf{Reasoning Content:}
  ``Let me start over with a cleaner approach.''

  \begingroup
  \setlength{\parskip}{0pt}
  \textbf{Command:}\par
  \begin{lstlisting}[style=trace]
  rm -rf /git /git/* 2>/dev/null
  mkdir -p /git/project /git/www/main /git/www/dev ...
  cd /git/project && git init --bare && ls -la
  \end{lstlisting}
  \textbf{Result:}\par
  \begin{lstlisting}[style=trace]
  Initialized empty Git repository in /git/project/
  \end{lstlisting}
  \endgroup

  \textbf{Turns 17 -- 22 (condensed):}

  \textbf{Main work:}
  The agent writes and revises the post-receive hook while working
  with the newly created repository.

  \textbf{Turn 23:}

  \begingroup
  \setlength{\parskip}{0pt}
  \textbf{Command:}\par
  \begin{lstlisting}[style=trace]
  rm -rf /git /git/* 2>/dev/null &&
  mkdir -p /git/project /git/www/main /git/www/dev
  \end{lstlisting}
  \endgroup

  \textbf{Turns 24 -- 42 (condensed):}

  \textbf{Main work:}
  The agent rebuilds the bare repository, hook, and Nginx
  configuration. The Nginx syntax check passes, but the agent
  stops before completing a Git push and checking the deployed
  pages. The inherited state therefore contains useful setup
  alongside unresolved SSH and deployment problems.
  \end{promptbox}

  \begin{promptbox}[redprompt]{GRPO -- Failed Recovery}
  \setlength{\parindent}{0pt}
  \setlength{\parskip}{0.6\baselineskip}

  \textbf{Turn 43:}

  \textbf{Reasoning Content:}
  ``Let me continue the setup. I need to ... test the full
  workflow.''

  The agent attempts to configure SSH access.

  \begingroup
  \setlength{\parskip}{0pt}
  \textbf{Result:}\par
  \begin{lstlisting}[style=trace]
  Need to setup SSH key
  \end{lstlisting}
  \endgroup

  \textbf{Turn 46:}

  \textbf{Reasoning Content:}
  ``I need to set up the directories correctly. Let me create
  a proper setup. ... Let me redesign.''

  \begingroup
  \setlength{\parskip}{0pt}
  \textbf{Command:}\par
  \begin{lstlisting}[style=trace]
  rm -rf /git /git/* 2>/dev/null
  mkdir -p /git/project /git/www /git/www_dev ...
  cd /git/project && git init --bare
  \end{lstlisting}
  \textbf{Result:}\par
  \begin{lstlisting}[style=trace]
  Initialized empty Git repository in /git/project/
  \end{lstlisting}
  \endgroup

  \textbf{Turn 48:}

  \textbf{Reasoning Content:}
  ``Let me clean up and start fresh with a simpler, more reliable
  approach.''

  \begingroup
  \setlength{\parskip}{0pt}
  \textbf{Command:}\par
  \begin{lstlisting}[style=trace]
  rm -rf /git /git/* /git/www /git/main /git/dev 2>/dev/null
  cd /git && git init --bare project.git 2>/dev/null || true
  \end{lstlisting}
  \textbf{Result:}\par
  \begin{lstlisting}[style=trace]
  bash: cd: /git: No such file or directory
  bash: /git/project.git/hooks/post-receive: No such file or directory
  chmod: cannot access '/git/project.git/hooks/post-receive':
  No such file or directory
  \end{lstlisting}
  \endgroup

  \textbf{Turns 49 -- 54 (condensed):}

  \textbf{Main work:}
  The agent repeatedly rebuilds the repository and hook, clearing /git several more times.

  \textbf{Turn 55:}

  \textbf{Reasoning Content:}
  ``Let me use a clearer, more standard approach for the
  post-receive hook.''

  \begingroup
  \setlength{\parskip}{0pt}
  \textbf{Command:}\par
  \begin{lstlisting}[style=trace]
  rm -rf /git /git/* /git/www /tmp/deploy* 2>/dev/null
  ...
  cat > /git/project/.git/hooks/post-receive << 'HOOKSCRIPT'
  \end{lstlisting}
  \textbf{Result:}\par
  \begin{lstlisting}[style=trace]
  Initialized empty Git repository in /git/project.git/
  bash: /git/project/.git/hooks/post-receive: No such file or directory
  chmod: cannot access '/git/project/.git/hooks/post-receive':
  No such file or directory
  Hook created
  ls: cannot access '/git/project/.git/hooks/post-receive':
  No such file or directory
  \end{lstlisting}
  \endgroup
  \end{promptbox}

  \begin{promptbox}[greenprompt]{ReTerm-GRPO -- Successful Recovery}
  \setlength{\parindent}{0pt}
  \setlength{\parskip}{0.6\baselineskip}

  \textbf{Turn 43:}

  \textbf{Reasoning Content:}
  ``Let me start by auditing the current state---what's been set up
  and what still needs to be done.''

  The agent inspects the inherited repository, hook, web
  directories, Nginx configuration, and running services.

  \textbf{Turn 44:}

  \textbf{Reasoning Content:}
  ``The post-receive exists but the hook deploys to both main and
  dev for EVERY branch push. This is incorrect. ...
  SSHD not running ... Nginx not running.''

  \textbf{Turns 45 -- 49 (condensed):}

  \textbf{Main work:}
  The agent revises the setup without deleting /git/project, then starts SSH.

  \textbf{Turn 50:}

  \textbf{Reasoning Content:}
  ``I need to create the /run/sshd directory and start sshd
  properly.''

  \begingroup
  \setlength{\parskip}{0pt}
  \textbf{Command:}\par
  \begin{lstlisting}[style=trace]
  mkdir -p /run/sshd && /usr/sbin/sshd -p 22
  sleep 1; ss -tlnp | grep 22
  \end{lstlisting}
  \textbf{Result:}\par
  \begin{lstlisting}[style=trace]
  LISTEN 0 128 0.0.0.0:22 0.0.0.0:* users:(("sshd",pid=10107,fd=3))
  \end{lstlisting}
  \endgroup

  \textbf{Turns 51 -- 54 (condensed):}

  \textbf{Main work:}
  The agent starts Nginx, checks the repository and hook, then tests SSH access.

  \textbf{Turn 55:}

  \begingroup
  \setlength{\parskip}{0pt}
  \textbf{Command:}\par
  \begin{lstlisting}[style=trace]
  sshpass -p "password" ssh -o "PasswordAuthentication=yes" \
    gituser@localhost "ls -la /git/project/"
  \end{lstlisting}
  \textbf{Result:}\par
  \begin{lstlisting}[style=trace]
  Permission denied, please try again.
  \end{lstlisting}
  \endgroup

  \textbf{Turn 57:}

  \textbf{Reasoning Content:}
  ``\emph{The password hash looks like it's locked (starts with !).}
  Let me re-set the password for gituser.''

  \begingroup
  \setlength{\parskip}{0pt}
  \textbf{Command:}\par
  \begin{lstlisting}[style=trace]
  echo 'gituser:password' | chpasswd &&
  sshpass -p "password" ssh -o "PasswordAuthentication=yes" \
    gituser@localhost "echo TEST_OK"
  \end{lstlisting}
  \textbf{Result:}\par
  \begin{lstlisting}[style=trace]
  TEST_OK
  \end{lstlisting}
  \endgroup
  \end{promptbox}

  \section{Baselines}
We compare our model with the following baselines:
GPT-5 mini \citep{singh2025openai},
Gemini 3 Flash \citep{google2025gemini3flash},
Claude Sonnet 4.5 \citep{anthropic2025sonnet45},
Claude Opus 4.6 \citep{anthropic2026opus46},
Gemini 3.1 Pro \citep{google2026gemini31pro},
GPT-5.5 \citep{openai2026gpt55},
GLM-4.7 \citep{zai2025glm47},
GLM-5 \citep{zeng2026glm},
Qwen3-8B, Qwen3-14B, and Qwen3-32B \citep{yang2025qwen3},
Qwen3.5-4B, Qwen3.5-9B, Qwen3.5-122B-A10B, and Qwen3.5-397B-A17B \citep{qwen3.5},
Kimi K2.5 \citep{team2026kimi},
MiniMax-M2.5 \citep{team2026minimax},
TMax-4B and TMax-9B \citep{ivison2026tmax},
Socratic-SWE \citep{xiao2026socratic},
ECHO-14B \citep{shrivastava2026echo},
Terminal-World-32B \citep{cheng2026terminal},
TermiGen-32B \citep{zhu2026termigen},
OpenThinkerAgent-32B \citep{raoof2026openthoughts},
and Nemotron-Terminal-32B \citep{pi2026data}.
\end{document}

%% file: math_commands.tex
\usepackage{amsmath,amsfonts,bm}

\def\eqref#1{equation~\ref{#1}}
\def\1{\bm{1}}

\DeclareMathAlphabet{\mathsfit}{\encodingdefault}{\sfdefault}{m}{sl}
\SetMathAlphabet{\mathsfit}{bold}{\encodingdefault}{\sfdefault}{bx}{n}